\documentclass[lettersize,journal]{IEEEtran}
\usepackage{amsmath,amsfonts}

\usepackage{array}
\usepackage[caption=false,font=normalsize,labelfont=sf,textfont=sf]{subfig}
\usepackage{textcomp}
\usepackage{stfloats}
\usepackage{url}
\usepackage{verbatim}
\usepackage{bm}
\usepackage{graphicx}
\usepackage{algorithm}
\usepackage{algpseudocode}
\usepackage{etoolbox,lipsum}
\usepackage[english]{babel}
\usepackage{makecell}

\usepackage[utf8]{inputenc}

\usepackage[backend=biber,style=ieee, doi=false, url=false, isbn=false, eprint=false]{biblatex}

\usepackage{tikz}
\usetikzlibrary{shapes, arrows, positioning}
\usetikzlibrary{arrows.meta, shapes.geometric, shapes.misc, fit, backgrounds, shadows.blur}

\usepackage[x11names]{xcolor}

\usepackage{fixmath}

\begin{document}

\title{ 
Learning Stack-of-Tasks Management for Redundant Robots
}

\author{Alessandro Adami$^{* \ddagger}$, Aris Synodinos$^\dagger$, Matteo Iovino$^{\dagger\mathsection}$, Ruggero Carli$^*$, Pietro Falco$^*$

\thanks{These experiments were carried out in the WASP Research Arena (WARA)-Robotics, hosted by ABB Corporate Research Center in Västerås, Sweden and financially supported by the Wallenberg AI, Autonomous Systems, and Software Program (WASP) funded by the Knut and Alice Wallenberg Foundation. 
\\Project co-funded by the European Union – Next Generation Eu - under the National Recovery and Resilience Plan (NRRP), Mission 4 Component 2, Investment 3.3 – Decree no. 630 ($24^{th}$ April 2024)  of Italian Ministry of University and Research; Concession Decree no. 1956 del $05^{th}$ December 2024 adopted by the Italian Ministry of University and Research, CUP D93D24000270003, within the national PhD Programme in Autonomous Systems (XL cycle).}
\thanks{\\$*$  University of Padova, Dept. of Information Engineering, Italy.\\
$\dagger$ABB Corporate Research, Västerås, Sweden.\\
$\mathsection$ Mobile Robotics Lab, ETH Zürich, Zürich, Switzerland. \\
$\ddagger$ Polytechnic of Bari Dept. of Electrical and Information Engineering, Italy.}
}

\markboth{Journal }%
{Shell}


\maketitle

\begin{abstract}
This paper presents a novel framework for automatically learning complete Stack-of-Tasks (SoT) controllers for redundant robotic systems, including task priorities, activation logic, and control parameters. Unlike classical SoT pipelines—where task hierarchies are manually defined and tuned—our approach optimizes the full SoT structure directly from a user-specified cost function encoding intuitive preferences such as safety, precision, manipulability, or execution speed. The method combines Genetic Programming with simulation-based evaluation to explore both discrete (priority order, task activation) and continuous (gains, trajectory durations) components of the controller. We validate the framework on a dual-arm mobile manipulator (the ABB mobile-YuMi research platform), demonstrating robust convergence across multiple cost definitions, automatic suppression of irrelevant tasks, and strong resilience to distractors. Learned SoTs exhibit expert-like hierarchical structure and adapt naturally to multi-objective trade-offs. Crucially, all controllers transfer from Gazebo simulation to the real robot, achieving safe and precise motion without additional tuning. Experiments in static and dynamic environments show reliable obstacle avoidance, high tracking accuracy, and predictable behavior in the presence of humans. The proposed method provides an interpretable and scalable alternative to manual SoT design, enabling rapid, user-driven generation of task execution hierarchies for complex robotic systems.
\end{abstract}


\section{Introduction} \label{introduction}
\begin{figure*}[t]
\centering
\begin{tikzpicture}[
    font=\sffamily,
    node distance=0.8cm and 1.6cm,
    arrow/.style={->, semithick, >=Latex},
    phase/.style={
        rectangle,
        rounded corners,
        thick,
        fill=blue!12,
        minimum width=5.6cm,
        minimum height=4.2cm,
        inner sep=5pt
    },
    block/.style={
        rectangle,
        rounded corners,
        thick,
        fill=white,
        align=center,
        minimum width=2.5cm,
        minimum height=1.0cm,
        inner sep=3pt
    },
    titlebar/.style={
        rectangle,
        rounded corners,
        fill=black!80,
        text=white,
        minimum width=5.6cm,
        minimum height=0.55cm,
        align=center,
        font=\bfseries\footnotesize
    }
]

\node[phase, minimum width=9.5cm, minimum height=6.0cm] (P1) {};
\node[titlebar, anchor=north] at (P1.north) {LEARNING FRAMEWORK};

\node[block, anchor=north] (sim)
    at ($(P1.north) + (-2.5,-0.75)$)
    {\includegraphics[width=2.0cm]{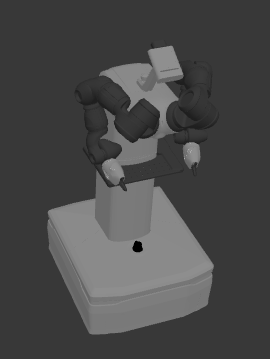}\\
     \textbf{Simulation}\\SoTs are executed};

\node[block, below=0.35cm of sim] (cost)
    {\textbf{Fitness Evaluation}\\$
    C = \sum_{i=1}^{M} \alpha_i\,\widetilde{c}$};

\node[phase, right=-4.0cm of P1, minimum width=3.5cm, minimum height=4.5cm, fill=gray!30] (P2) {};
\node[titlebar, anchor=north, minimum width=2.5cm] at (P2.north) {GENETIC\\ PROGRAMMING};

\node[block, anchor=north] (sel)
    at ($(P2.north) + (0,-0.95)$)
    {\textbf{Selection}\\High-fitness SoTs};

\node[block, below=0.55cm of sel] (rec)
    {\textbf{Genetic} \\\textbf{Recombination}:
    \\mutation \&\\crossover};
\draw[arrow] (sel) -- (rec);
\node[phase, right=1.7cm of P2, minimum width=6.5cm, minimum height=6cm] (P3) {};
\node[titlebar, anchor=north] at (P3.north) {LABORATORY TESTS};

\node[block, anchor=north] (test)
    at ($(P3.north) + (0,-0.75)$)
    {\includegraphics[width=2.0cm]{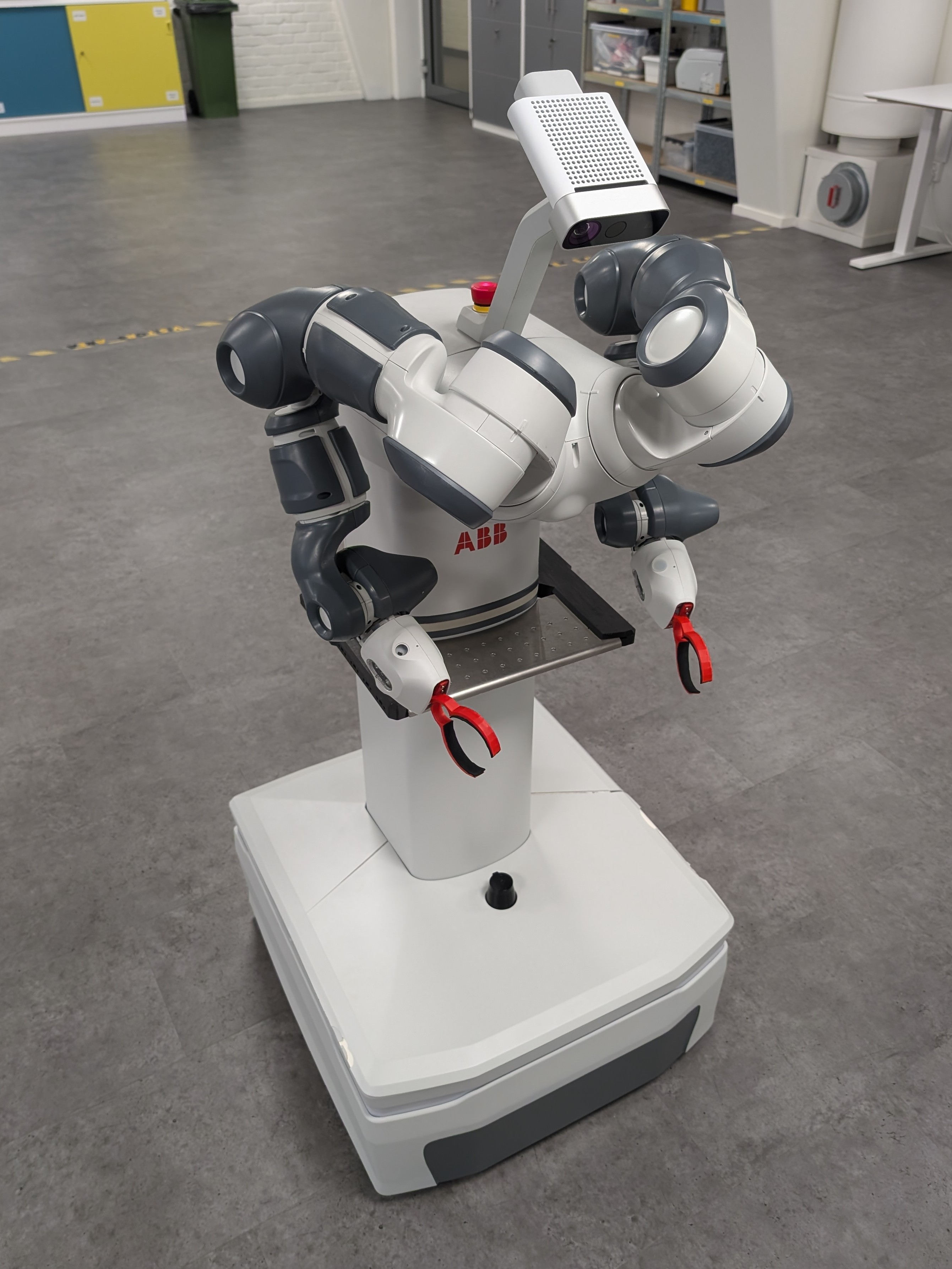}\\
     Real-Robot Execution};

\node[block, below=0.75cm of test] (val)
    {Performance Validation};
\draw[arrow] (test) -- (val);
\tikzset{bigarrow/.style={-latex, thick, line width=2pt}}
\draw[arrow] (sim) -- (cost);
\draw[bigarrow] (P1.east) --  (P3.west);
\draw[bigarrow] (cost.east) -| ++(0.3,0) -| ++(0.0, 0.4) -- ++(1.3, 0.0);
\draw[bigarrow] (P2.west) -- ++(-1.6, 0.0);
\end{tikzpicture}

\caption{Overview of the proposed framework. Candidate Stack-of-Tasks (SoT) controllers are first evaluated in simulation to compute a task-specific cost. A Genetic Programming process selects and recombines promising SoTs to iteratively generate improved candidates. The best evolved SoT is subsequently deployed on the ABB YuMi mobile manipulator for experimental validation.}
\label{fig:overall}
\end{figure*}

\IEEEPARstart{M}{odern} robotic systems increasingly operate in complex and dynamic environments where traditional, hand-crafted control architectures often lack the flexibility needed for reliable performance. Mobile manipulators---such as the mobile-YuMi research platform---combine high dexterity with mobility, resulting in significant kinematic redundancy. This redundancy enables the robot to pursue multiple objectives (e.g., obstacle avoidance, manipulability maximization, or joint-limit safety~\cite{MultipleTaskPriority}) but also complicates decision-making, especially when several tasks must be executed concurrently. Existing hierarchical control frameworks, such as~\cite{HierarchicalRedundancyResolutionUnderArbitraryConstraints,PickAndPlaceYumi,LowLovelFlexiblePlanning}, address these challenges through manually designed Stacks of Tasks (SoTs), requiring expert knowledge and extensive reprogramming when mission goals or environments change.
To overcome these limitations, we propose a framework (schematized in Fig.~\ref{fig:overall}) that \emph{automatically} learns both the structure and parameters of an SoT from a high-level cost function, eliminating the need for manual task prioritization. The problem addressed in this paper is to learn a hierarchy of prioritized tasks—i.e., ordering, activation, and control parameters—that optimizes a user-defined cost function. Formally, we seek to minimize $C(P,\theta)$ over discrete priority permutations P and continuous parameter vectors $\theta$, where the dimensionality and mixed nature of this space make classical optimization intractable.
\\The approach uses Genetic Programming (GP)~\cite{GPTutorial} to evolve candidate SoT configurations and employs reinforcement-learning-style episodic evaluation to select those that minimize a user-defined cost. GP naturally handles variable-length, hierarchical representations~\cite{GAandGPComparison}, making it well suited for discovering task sequences, control gains, and activation flags —elements that are, in part, inherently discrete and non-differentiable.
The resulting framework enables robots to autonomously construct SoTs that are aligned with user-defined preferences on precision, safety, manipulability, and execution time. The learned task hierarchies are independent of specific environments, and parameters are learned kinematically, allowing direct sim-to-real transfer without manual retuning. Furthermore, training the SoTs in Gazebo simulation will give accurate enough behaviors, compared to real-world scenarios.
We validate this methodology in both Gazebo simulations and real experiments on the mobile-YuMi platform, demonstrating adaptive behavior in scenarios involving static and moving obstacles. This work contributes to the development of adaptive, human-aligned control architectures (described in Sec.~\ref{sec: comparison}) for complex mechatronic systems, addressing autonomy in high-DOF robotic platforms.
The main contributions of this work are twofold:
\begin{itemize}
    \item A method for automatically learning SoT structures---including task order, parameters, and activation---using GP as a policy-search mechanism informed by episodic cost.
    \item A fully automated redundancy-management pipeline that generalizes across environments and requires no manual SoT engineering, enabling scalable and adaptive robot behavior. The approach is validated through extensive simulation and real-world experiments.
\end{itemize}

\noindent
To the best of our knowledge, no prior work jointly learns task priorities, activation logic, and control parameters within a Stack-of-Tasks framework. We benchmark the proposed optimization framework against both manually designed SoTs and heuristically generated ones produced by a Large Language Model from high-level task and cost descriptions.
\\Future works will explore the opportunity to integrate Gradient Descent algorithms in the non-differentiable and non-linear proposed SoT structure.
\\The remainder of this paper is organized as follows: Section~\ref{Related Works} reviews related work; Sections~\ref{sec:problem}--\ref{Theoretical framework details} present the proposed framework; Section~\ref{Experimental setup} describes the simulation environment and experimental platform; Section~\ref{Experimental results} reports the results; and Section~\ref{Conclusion and future works} concludes the paper.

\section{Related Works} \label{Related Works}

\subsection{Classical Stack-of-Tasks Framework}
The Stack-of-Tasks (SoT) framework is a well-established approach for redundancy resolution and task prioritization in robotics. Classical SoT implementations manually define both the task objectives and their hierarchy. For instance,~\cite{SoTInvertedKin} presents a generalized SoT formulation for humanoid control, while~\cite{SoTDynMotion} applies SoT principles to motion capture and re-targeting. Extensions to the SoT framework include the incorporation of unilateral constraints~\cite{SoTUnilateralConstarints} and visuo--tactile feedback~\cite{SoTVisuoTactile}. More recently, SoT has been combined with behavior Trees~\cite{SoTWithBT}, improving modularity and interpretability. However, these systems still rely on manually crafted task definitions and fixed priority structures.

\subsection{Optimization-Based and Hybrid Hierarchical Control}
Beyond classical SoTs, several optimization-based and hybrid methods have been proposed for redundancy resolution and multi-objective robot control. Genetic Algorithms (GA)~\cite{GAReview} and Genetic Programming (GP) have been explored for motion generation and planning, including collision-aware path planning~\cite{GARedundantManip} and inverse-kinematics optimization~\cite{IKRedRobotsGA}. Other hierarchical control schemes, such as those based on null-space projections~\cite{LowLovelFlexiblePlanning} or behavior Trees~\cite{LearningBTwithGPUmpredictEnv}, enable modular task composition but still require explicit developer-defined task structures. Hybrid planning approaches, such as the two-layer sampling and set-based IK strategy in~\cite{MerginGlobalAndLocalPlanners}, improve reactivity in dynamic environments, though they depend strongly on accurate perception and do not scale well to robots with many degrees of freedom.

\subsection{Learning-Based Control and Its Limitations}
Deep Reinforcement Learning~\cite{RLSutton1998,RLSurveyShakya} offers powerful tools for decision-making in high-dimensional robotic settings~\cite{RLinRoboticsSurvey}. However, RL policies typically encode task priorities implicitly within a neural controller, making the resulting behaviors difficult to interpret, tune, or modify when safety or precision requirements change. Crucially, RL-based methods do not produce explicit task hierarchies, nor do they guarantee that lower-priority objectives remain subordinate to safety-critical constraints. Moreover, none of these learning-based strategies attempts to derive a full SoT structure—including task ordering, activation logic, and control parameters—with explicit null-space projections.

\subsection{Our Contribution}
In most state-of-the-art robotic systems, redundancy management still requires expert designers to handcraft the hierarchy~\cite{LowLovelFlexiblePlanning}, which becomes increasingly impractical for highly redundant platforms such as dual-arm mobile manipulators~\cite{SoTWithBT}. In contrast, the approach proposed in this work \emph{automatically} learns both the structure and parameters of a SoT directly from a user-defined cost function. To the best of our knowledge, no prior method jointly learns the task order, activation flags, and control gains within a complete SoT architecture. By treating the SoT as a structured policy and evolving it via Genetic Programming, the proposed method enables autonomous redundancy management with direct sim-to-real transfer, without requiring expert tuning.

\section{Problem Statement and Proposed Solution}
\label{sec:problem}

\subsection{Problem Statement}
Consider a robot with configuration $\bm{q}\in\mathbb{R}^n$ executing tasks defined in an $m$-dimensional task space ($n>m$), where
\begin{equation}
    \bm{x} = \bm{k}(\bm{q}), \qquad 
\dot{\bm{x}} = \mathrm{J}(\bm{q})\,\dot{\bm{q}},
\end{equation}
and $\bm{k}$ is smooth with Jacobian $\mathrm{J}(\bm{q})$.
Redundancy ($n>m$) allows satisfying secondary objectives such as obstacle avoidance, manipulability maximization, or joint-limit safety.  \\The challenge is to determine which combination of such objectives best fulfills a high-level mission without manual engineering of the task
hierarchy.
\\Let $D=\{T_1,\dots,T_k\}$ be a dictionary of available tasks.
A Stack of Tasks (SoT) is defined as:
\begin{equation}
\mathrm{SoT}=(P,\theta,A),
\end{equation}
where $P\in\mathcal{P}_k$ is a priority ordering, $\theta\in\Theta^k$ are task parameters (e.g., gains), and $A\in\{0,1\}^k$ are activation flags. Given an initial configuration $\bm{q}_0\in\mathbb{R}^n$, an SoT induces a roll-out trajectory $q(P,\theta,A)$ via recursive null-space projection of the task velocities as $q_{next}=q_{current}+\int_{0}^{t}\dot{q}(P,\theta,A,t)\:dt$.
\\The user provides a cost function
\begin{equation}
C(P,\theta,A):q\in\mathbb{R}^n \mapsto \mathbb{R}_{\ge 0},
\end{equation}
expressing mission requirements (precision, safety, time, etc.). The learning objective is therefore:
\begin{equation}
\label{eq:mainopt}
(P^\ast,\theta^\ast,A^\ast)
= \underset{P,\theta,A}{\arg\min}\;
C(P,\theta,A),
\end{equation}
subject to:
\begin{align*}
    &\theta_i \in \Theta_i \quad \text{(stability constraints)},\\
    &q(P,\theta,A)\ \text{terminates in finite time},\\
    &\mathrm{SoT} \text{ is feasible under null-space semantics}.
\end{align*}
This formulation highlights that the search space is mixed discrete–continuous and highly nonconvex:  $P$ has $k!$ permutations, $A$ has $2^k$ possible activation patterns, and $\theta$ lies in a continuous bounded domain. Furthermore, the cost function $C$ is defined over $q$, potentially hiding the relation with task parameters and null-space projection over priorities.
Therefore, classical gradient-based optimization is not suitable.

\subsection{Proposed Solution}

To address~\eqref{eq:mainopt}, we propose an evolutionary framework that treats
the SoT as a structured policy and searches over 
$(P,\theta,A)$ using Genetic Programming (GP).  
The method is illustrated in Fig.~\ref{fig:overall} and summarized in Algorithm~\ref{alg: alg 1}.
\\An initial population of $n_T$ SoTs is generated by randomly sampling
$P$, $\theta$, and $A$ from their respective domains.
Each SoT is executed in simulation under randomized initial conditions until
success, failure (e.g., collision or singularity), or time-out.
Its episodic cost $C$ is then computed.
\\GP iteratively improves the population via:
\begin{itemize}
    \item \textbf{Tournament selection}~\cite{GPTournamentSelection}, retaining lower-cost individuals,
    \item \textbf{Crossover}~\cite{GPTutorial}, exchanging task elements between two parents,
    \item \textbf{Mutation}~\cite{GPTutorial}, modifying task parameters, priorities, or activation flags.
\end{itemize}
Activation flags implement an intron-based mechanism
\cite{GPIntronsBenefit,GPExplicitelyDefinedIntrons}, allowing the representation to explore structural variations without discarding useful genetic material. The loop continues until convergence, or until the population's SoTs are deemed equivalent according to the distance metric in Sec.~\ref{Sec: SoT dist}.
\\The best-performing SoT of the final population is then validated on the physical platform.
\begin{algorithm}[]
    \caption{Pseudo-code of the proposed solution}
    \begin{algorithmic}
        \State User defines cost function $C$
        \State Generate random initial SoT population $pop$
        \While{stop condition not met}
            \State Execute each SoT in $pop$ and compute its cost
            \State Apply tournament selection to retain the best individuals
            \State $new\_pop \gets \emptyset$
            \For{each survived $SoT$}
                \State $new\_pop \gets SoT$
                \State Select operation $gen\_op \in \{crossover, mutation\}$
                \If{$gen\_op$ is crossover}
                    \State Choose second parent, produce offspring $SoT'$
                \Else
                    \State Mutate $SoT$ to obtain $SoT'$
                \EndIf
                \State $new\_pop \gets SoT'$
            \EndFor
            \State $pop \gets new\_pop$
        \EndWhile
        \State \textbf{return} best SoT in $pop$
    \end{algorithmic}
    \label{alg: alg 1}
\end{algorithm}

\section{Theoretical Framework Details}
\label{Theoretical framework details}

This section formalizes the components used by the proposed learning framework:  task models, reference-generation schemes, null-space prioritization, the cost function, the Stack of Tasks representation, and the GP-based optimization operators.

\subsection{Tasks}
The robot executes tasks drawn from a predefined dictionary $D=\{T_1,\dots,T_n\}$. Each task $T_i$ is characterized by a task variable  $\bm{x}_i \in \mathbb{R}^{m_i}$ and a smooth task map $\bm{k}_i : \mathbb{R}^n \rightarrow \mathbb{R}^{m_i}$, with
\begin{equation}
\bm{x}_i = \bm{k}_i(\bm{q}), 
\qquad
\dot{\bm{x}}_i = \mathrm{J}_i(\bm{q})\,\dot{\bm{q}},
\end{equation}
where $\mathrm{J}_i(\bm{q})$ is the task Jacobian.  
Given a desired trajectory $\bm{x}_{i,d}(t)$, the controller must generate joint velocities $\dot{\bm{q}}_i(t)$ that reduce the cost.
Two standard reference-generation schemes are used:
\begin{itemize}
    \item \textbf{Closed-loop (tracking) tasks}~\cite{antonelli2008null,StabilityClosedLoopIK}:
    \begin{equation}
        \dot{\bm{q}}_i = 
        \mathrm{J}_i^{\dagger}(\bm{q})
        \big(\dot{\bm{x}}_{i,d} + 
        \gamma_{i}\big(\bm{x}_{i,d}-\bm{k}_i(\bm{q})\big)\big),
    \end{equation}
    where $\gamma_i>0$ is a feedback gain ensuring exponential tracking under standard regularity conditions.

    \item \textbf{Open-loop (gradient) tasks}~\cite{RoboticsSiciliano}:
    For scalar objectives $x_i(\bm{q})$ (e.g., manipulability or joint-limit avoidance),
    \begin{equation}
        \dot{\bm{q}}_i = \gamma_i \left(\nabla_{\bm{q}} x_i(\bm{q})\right)^T.
    \end{equation}
\end{itemize}
Each task therefore has a parameter vector $\theta_i$, constrained to a feasible domain $\Theta_i = [\theta_i^{\min},\theta_i^{\max}]$.

\subsection{Null-Space Prioritization}
Given two tasks with references $\dot{\bm{q}}_1$ and $\dot{\bm{q}}_2$, the hierarchical composition follows the classical null-space structure 
\cite{MultipleTaskPriority,RoboticsSiciliano}:
\begin{equation}
\dot{\bm{q}} = 
\dot{\bm{q}}_1 + 
\big(\bm{I}_n - \mathrm{J}_1^{\dagger}\mathrm{J}_1\big)\dot{\bm{q}}_2 = \dot{\bm{q}}_1 + N_1\dot{\bm{q}}_2.
\end{equation}
Thus, task~1 is executed exactly when feasible, while task~2 acts only in the null space of task~1. A priority-ordered sequence is constructed recursively for any $P\in\mathcal{P}_k$. We assume all Jacobians are full row rank or damped when near singularities.

\subsection{Cost Function}
\label{Cost}
A user-defined episodic cost function assigns a scalar evaluation to the trajectory induced by a SoT:
\begin{equation}
    C = \sum_{i=1}^{M} \alpha_i\,\widetilde{c}_i, 
    \qquad 
    \sum_i \alpha_i = 1,
    \label{eq: cost}
\end{equation}
where $c_i$ are individual cost terms (e.g., time, safety, precision), and $\widetilde{c}_i$ are normalized versions ensuring comparable numeric scales. Convex weighting maintains interpretability and prevents numerically dominant cost terms from biasing the search~\cite{poli2008field}.
\\This yields a bounded and smooth episodic landscape that improves GP convergence~\cite{bongard2013evolutionary}.
\\Weights can be adjusted via a GUI by a human user, basing them on specific requirements (Sec.~\ref{GUI}), or generated using language-based tools~\cite{TowardsAutonomousRLwithLLMs}.

\subsection{Stack of Tasks Representation}
A Stack of Tasks is encoded as the tuple
\begin{equation}
\mathrm{SoT} = (P,\theta,A),
\end{equation}
where $P$ is a priority ordering, $\theta = (\theta_1,\dots,\theta_m)$ are task parameters, and $A\in\{0,1\}_k$ are activation flags. Inactive tasks ($A_i=0$) do not influence the control law but remain in the genotype, following intron-based representations known to improve GP robustness
\cite{GPIntronsBenefit,GPExplicitelyDefinedIntrons}.
\\Each individual in the GP population is one complete SoT. Boundary constraints on $\theta_i$ enforce feasibility (e.g., $\gamma_i < 2$ for closed-loop stability \cite{StabilityClosedLoopIK}).
\\A candidate SoT is rolled out until:
\begin{itemize}
    \item the mission is completed,
    \item an unsafe condition occurs (collision or singularity),
    \item or a time limit is reached.
\end{itemize}
The induced trajectory, computed as in Fig.~\ref{fig:compact_sot_three_inputs}, yields its episodic cost $C$.
\begin{figure}[t]
\centering
\begin{tikzpicture}[
    task/.style={
        rectangle,
        rounded corners,
        minimum width=4.2cm,
        minimum height=0.75cm,
        thick,
        fill=gray!10,
        font=\small,
        align=left
    },
    arrow/.style={->, thick},
    node distance=0.15cm
]

\node[task] (T1)
    {\textbf{Task 1 (priority 1)}\\
     $\dot{\mathbf{q}}_{1} = \mathrm{J}_1^\dagger \mathbf{y}_1,\:N_1$};

\node[task,below=of T1] (T2)
    {\textbf{Task 2 (priority 2)}\\
     ${\mathbf{q}}_{2} = \mathrm{J}_2^\dagger \mathbf{y}_2,\:N_{12}$};

\node[task,below=of T2] (T3)
    {\textbf{Task 3 (priority 3)}\\
     ${\mathbf{q}}_{3} = \mathrm{J}_3^\dagger \mathbf{y}_3$};


\node[font=\small] (E1) [right=1.6cm of T1] {$\dot{\mathbf{q}}_{1}$};
\node[font=\small] (E2) [right=1.6cm of T2] {$+\, N_{1}\dot{\mathbf{q}}_{2}$};
\node[font=\small] (E3) [right=1.6cm of T3] {$+\, N_{12}\dot{\mathbf{q}}_{3}$};

\node[font=\small, below=0.3cm of E3, align=center] (EQ) 
{
$\displaystyle
\dot{\mathbf{q}}
= 
\dot{\mathbf{q}}_{1}
+ N_{1}\dot{\mathbf{q}}_{2}
+ N_{12}\dot{\mathbf{q}}_{3}
$
};


\draw[arrow] (T1.east) -- (E1.west);
\draw[arrow] (T2.east) -- (E2.west);
\draw[arrow] (T3.east) -- (E3.west);

\draw[arrow] (T1.east) -| ++(0.3,0) -| ++(0.0, -0.4) -| ++(1.85, 0.0) -| ++ (0.0, -0.4);
\draw[arrow] (T2.east) -| ++(0.3,0) -| ++(0.0, -0.4) -| ++(1.85, 0.0) -| ++ (0.0, -0.4);
\draw[arrow] (E3.south)++(-0.2, 0.0) -- ++(0.0, -0.3);
\end{tikzpicture}
\caption{Schematic of a three-task Stack of Tasks (SoT) null space projection. Each task contributes 
to the final joint velocity command through its own projected reference, forming 
the hierarchical control law 
$\dot{\mathbf{q}}=\dot{\mathbf{q}}_{1}+N_1\dot{\mathbf{q}}_{2}+N_{12}\dot{\mathbf{q}}_{3}$.}
\label{fig:compact_sot_three_inputs}
\end{figure}

\subsection{Distance Between Stacks of Tasks}
\label{Sec: SoT dist}
A distance metric is used to detect structurally similar individuals and prevent iterating over a stationary population.  
For two SoTs $a$ and $b$ of equal length~$k$:
\begin{equation}
    d(a,b)=\sum_{i=1}^{k}
    \left(
    \delta_{\theta}(\theta^a_i,\theta^b_{j(i)}) +
    \delta_{\text{prior}}(T^a_i,T^b_{j(i)})
    \right),
\end{equation}
where $k(i)$ is the position in $b$ of the same task appearing in position $i$ of~$a$. The terms compare numerical parameters and priority structure:
\[
\delta_{\theta}(\theta^a_i,\theta^b_{j(i)}) 
= \|\theta^a_i - \theta^b_{j(i)}\|^2,
\]
\begin{equation}
\delta_{\text{prior}}(T^a_i,T^b_{j(i)}) =
\begin{cases}
|\pi^a_i - \pi^b_{j(i)}|, & \text{if both tasks active},\\
n, & \text{otherwise},
\end{cases}
\end{equation}
where $\pi^a_i$ denotes the priority index (among active tasks) of task $i$ in $a$. This metric identifies SoTs that are simultaneously similar in terms of parameters and hierarchical structure.

\subsection{Genetic Programming as a Cost Minimizer}
Genetic Programming (GP)~\cite{GPTutorial} evolves structured programs, making it well-suited for optimizing the hierarchical SoT representation.  
In our setting, each SoT is a policy, the simulator acts as the environment, and the episodic cost is analogous to a reward signal or fitness function.
\\Variation operators include:
\begin{itemize}
    \item \textbf{Crossover} (binary operator):
    \begin{equation}
        \mathrm{SoT}' = \text{crossover}(\mathrm{SoT}_1,\mathrm{SoT}_2),
    \end{equation}
    which exchanges task elements or substructures between parents to produce a new offspring.

    \item \textbf{Mutation} (unary operator):
    \begin{equation}
        \mathrm{SoT}' = \text{mutation}(\mathrm{SoT}),
    \end{equation}
    altering parameters, priorities, or activation flags to produce a new offspring.
\end{itemize}
These operators maintain structural diversity and avoid premature convergence, allowing exploration~\cite{GPandMLSurvey,GraphRepresentationsGP,LearningBTCollaborativeRobot}.
\\Binary \textbf{tournament selection} chooses the better of two uniformly sampled SoTs:
\begin{equation}
S^\star = 
\begin{cases}
S_1, & C(S_1) < C(S_2),\\[4pt]
S_2, & \text{otherwise}.
\end{cases}
\end{equation}
This deterministic scheme guarantees that lower-cost individuals survive, guiding the population toward solutions with reduced episodic cost. Discarded SoTs are lost forever from the algorithmic pipeline.

\section{Experimental Setup}\label{Experimental setup}

This section describes the simulation-based learning phase and the real-world validation on the mobile-YuMi research platform at ABB Corporate Research (Västerås, Sweden), together with its Gazebo digital twin. The robot consists of an omnidirectional mobile base and two 7-DOF arms. 
Only the base is equipped with a LiDAR sensor; therefore, obstacle avoidance affects base motion exclusively. Both simulation and real-robot experiments operate at the kinematic level, using joint velocities for the arms and planar velocities for the base. Furthermore, simulations are performed in a high-fidelity dynamics simulation environment.
This ensures that SoT structures learned in simulation can be transferred to the real robot. All interfaces are implemented in ROS~2 Humble.

\subsection{Task Set}
All tasks belong to a predefined dictionary, and return desired velocity commands $\dot{\bm{q}}_d\in\mathbb{R}^{n}$, consistent with the general task model defined in Sec.~\ref{Theoretical framework details}. Tasks are grouped into: \emph{mission-relevant tasks} used by the robot for goal execution, and  \emph{non-relevant tasks} deliberately included to test robustness against distracting behaviors.

\begin{figure}[H]
            \centering
            \includegraphics[width=0.24\linewidth]{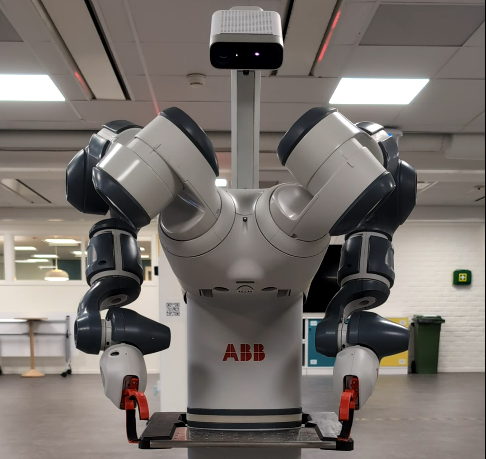}
            \includegraphics[width=0.24\linewidth]{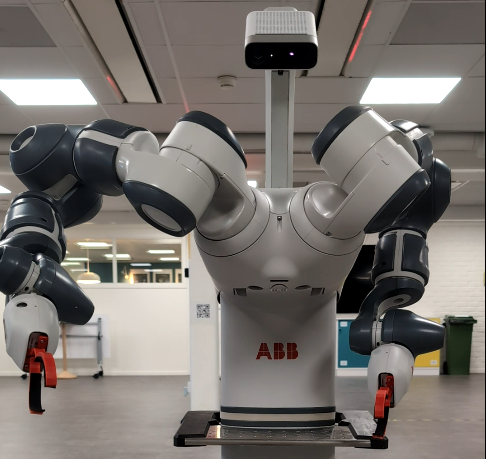}
            \includegraphics[width=0.24\linewidth]{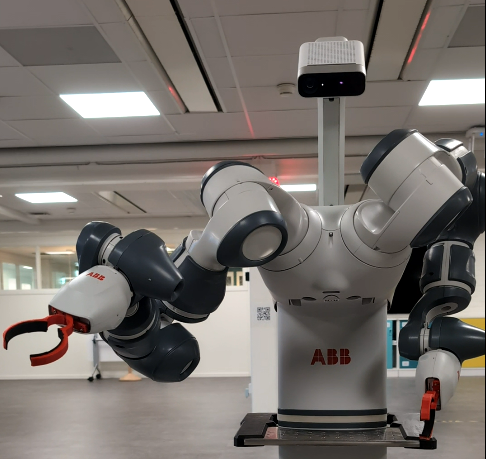}
            \includegraphics[width=0.24\linewidth]{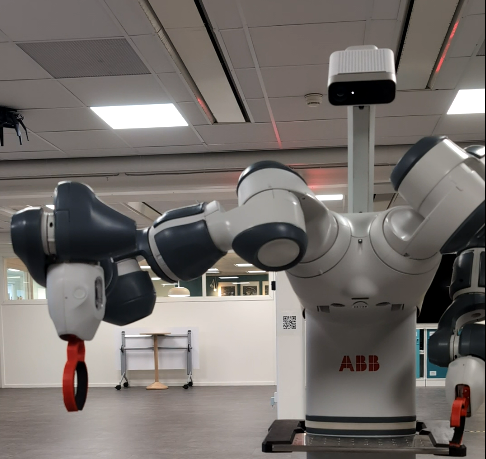}
            \caption{Example of the robot performing an Inverse Kinematic sequence.}
            \label{fig: IK}
        \end{figure}

\paragraph{Inverse Kinematics (IK)}
A time-parametrized Cartesian trajectory $\bm{k}_d(t)$ is generated from the mission specification. At each control step,
\begin{equation}
    \dot{\bm{q}}_d(t)=\mathrm{J}^{\dagger}(\bm{q}(t))
    \left(\dot{\bm{k}}_d(t)
    +\gamma_{CL}\big(\bm{k}_d(t)-\bm{k}(\bm{q}(t))\big)\right),
\end{equation}
where $\gamma_{CL}$ enforces closed-loop Cartesian tracking. Visual example of task execution i reported in Fig~\ref{fig: IK}.
\paragraph{Obstacle Avoidance (OA)~\cite{LowLovelFlexiblePlanning}}
The LiDAR scan is downsampled to a $180^\circ$ sector and modeled as $l$ virtual springs of rest length $r_l$. For each spring, a pseudo-energy $\epsilon_l=\frac12(r_l-d)^2$ is generated when $d<r_l$. The aggregated variable $\bm{\sigma}=\sum_l \epsilon_l$ is tracked through
\begin{equation}
    \dot{\bm{q}}_d(t) =
    \mathrm{J}_o^\dagger
    \left(\dot{\bm{\sigma}}_d(t)
    + \gamma_{CL}\big(\bm{\sigma}_d(t)-\bm{\sigma}(\bm{q}(t))\big)\right).
\end{equation}
If no obstacles are detected, $\mathrm{J}_o$ is empty and the task becomes inactive, allowing lower-priority tasks to determine the motion.

\paragraph{Manipulability Maximization}
Manipulability is computed as
\begin{equation}
    w(\bm{q})=\sqrt{\det\!\left(\mathrm{J}(\bm{q})\,\mathrm{J}^T(\bm{q})\right)}.
\end{equation}
The open-loop gradient controller is
\begin{equation}
    \dot{\bm{q}}_d=\gamma_{OL}
    \left(\frac{\partial w(\bm{q})}{\partial\bm{q}}\right)^{\!T}.
\end{equation}

\paragraph{Joint-limit Avoidance}
A convex task prevents proximity to mechanical limits:
\begin{equation}
    w(\bm{q})=-\frac{1}{2n}\sum_{i=1}^n
    \left(
        \frac{q_i-\bar{q}}
             {q_{i_{\max}}-q_{i_{\min}}}
    \right)^{\!2},
\end{equation}
whose gradient is used in the same form as above. For future work, it would be beneficial to incorporate self-collision checks, since joint limits alone do not always prevent self-collisions; for instance, a link of the arm may collide with the torso while performing a task.

\paragraph{Non-relevant Tasks}
These tasks generate circular base motion, pure yaw rotation, end-effector oscillations, or sinusoidal joint motions. Their Jacobians are intentionally incompatible with the mission objectives, creating perturbations that challenge the resilience of the evolved SoT hierarchy.

\subsection{Cost Definitions}
All cost components are normalized and combined into the convex form of Eq.~\eqref{eq: cost}. The following metrics are used:

\begin{itemize}
    \item \textbf{Precision:} squared pose error 
    $\widetilde{c}_i=\|\bm{k}_d-\bm{k}\|^2$.
    \item \textbf{Safety:} pseudo-energy derived from distances $d<r_l$,
    \[
        \widetilde{c}_i =
        \sum_{l=1}^m 
        \begin{cases}
            \frac{1}{2}(d-r_l)^2, & d < r_l,\\[2pt]
            0, & \text{otherwise},
        \end{cases}
    \]
    or equivalently using the minimum allowed distance $r_{\min}$.
    \item \textbf{Manipulability:} inverse squared measure
    $\widetilde{c}_i = 1/(w_{\text{max.manip}}^2+\epsilon)$,
    penalizing low manipulability.
    \item \textbf{Joint-limit proximity:} 
    $\widetilde{c}_i = w_{\text{MJL}}^2$.
    \item \textbf{Time:} 
    $\widetilde{c}_i = t^2$,
    penalizing slow task completion.
\end{itemize}

\subsection{Graphical User Interface (GUI)}\label{GUI}
\begin{figure}
    \centering
    \includegraphics[width=0.51\linewidth]{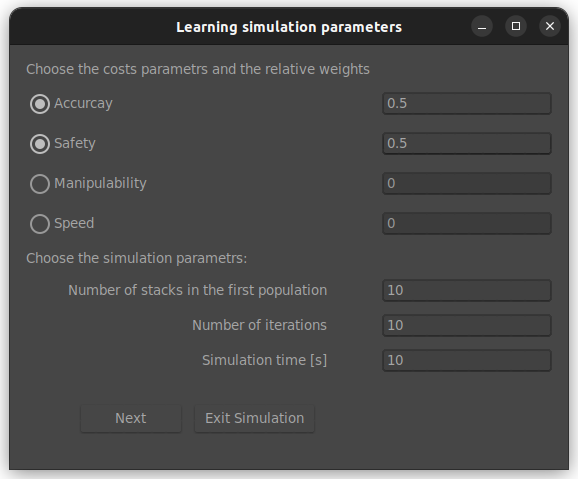}
    \includegraphics[width=0.46\linewidth]{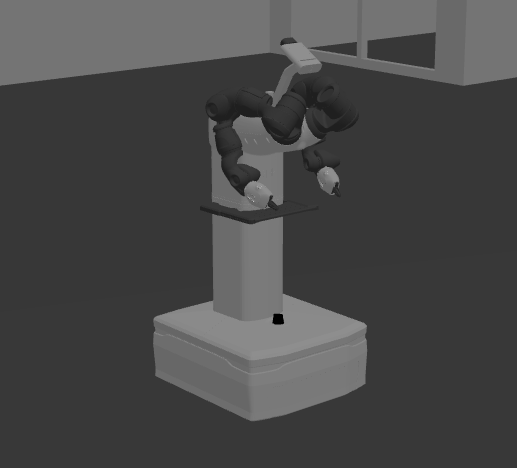}
    \caption{Graphical User Interface (left) for configuring learning simulation (right) parameters. The interface allows users to define the relative weights of cost criteria (Accuracy, Safety, Manipulability, Speed) and set simulation parameters, including population size, number of iterations, and simulation duration. Designed for ease of use, it supports intuitive navigation with dedicated controls for starting or exiting the simulation.}
    \label{fig:GUI}
\end{figure}

A lightweight GUI (Fig.~\ref{fig:GUI}) allows users to adjust cost weights and simulation parameters (population size, iteration count, and episode duration). This enables non-expert operators to configure learning experiments without interacting with source code.
\subsection{Gazebo Simulation and Learning Pipeline}
Simulation is performed in a high-fidelity Gazebo twin of the WARA Robotics Lab at ABB Corporate Research Center in Västerås, Sweden. At the start of each learning run, a population of SoT candidates is initialized randomly within safe gain ranges. Genetic Programming iteratively evolves this population, evaluating each SoT under randomized initial arm configurations and obstacle placements.  
\\The best-performing SoT of the final population is selected for real-robot testing. Joint velocities are saturated near mechanical limits according to
\begin{equation}
    \epsilon\,(q_{i_{\min}}-q_i)
    \le
    \dot{q}_i
    \le
    \epsilon\,(q_{i_{\max}}-q_i),
\end{equation}
ensuring safe operation in both simulation and hardware.  
Both static and dynamic obstacle scenarios are included, and LiDAR offsets are compensated for precise obstacle mapping.

\subsection{Real-Robot Experiments}
Learned SoTs are deployed on the physical mobile-YuMi through ROS~2, maintaining the same control interface as in simulation. Tests are performed under:
\begin{itemize}
    \item static and dynamic base-only conditions,
    \item static arm-only tasks,
    \item combined base–arm missions with static and dynamic obstacles.
\end{itemize}
Dynamic scenes include human motion near the robot. Standard ABB safety layers and manual emergency stops remain active throughout testing. All reported results use the same SoT learned entirely in simulation without manual fine-tuning.

\section{Experimental results}\label{Experimental results}

This section presents a detailed evaluation of the learning pipeline, the behavior of the learned Stacks of Tasks (SoTs), and the results obtained in real-world tests. The goals of the evaluation are to (i) assess convergence properties during learning, (ii) quantify robustness to irrelevant tasks and dynamic perturbations, and (iii) verify that the learned configurations transfer reliably from simulation to the physical robot.

\subsection{Learning of the Stack of Tasks}

The learning phase is divided into two sub-problems: discovering the priority order and optimizing the task parameters to show convergence and reliability of both scenarios. However, bot optimizations can be performed together in a longer, fully automated learning phase. In what follows, we report qualitative and quantitative behaviors observed across multiple runs.

\subsubsection{Priority order learning}

\begin{figure}[th]
        \centering
        \includegraphics[width=0.49\linewidth]{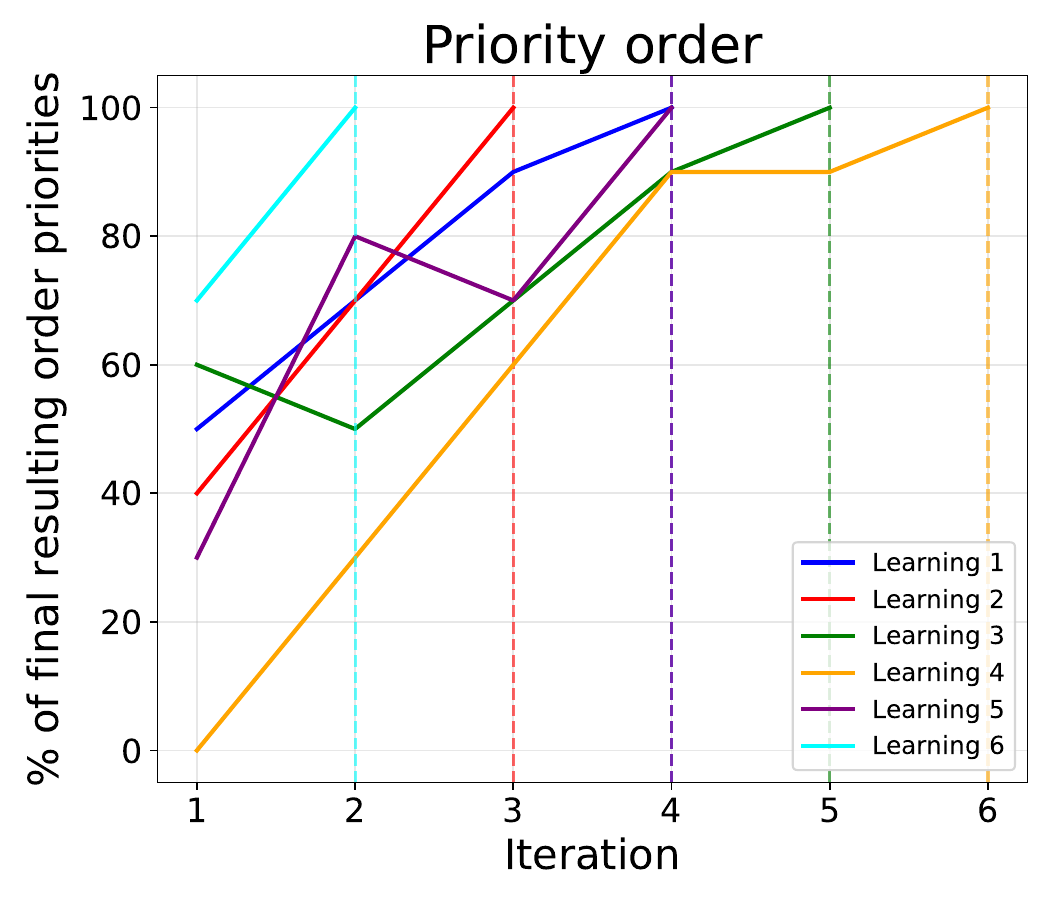}
        \includegraphics[width=0.49\linewidth]{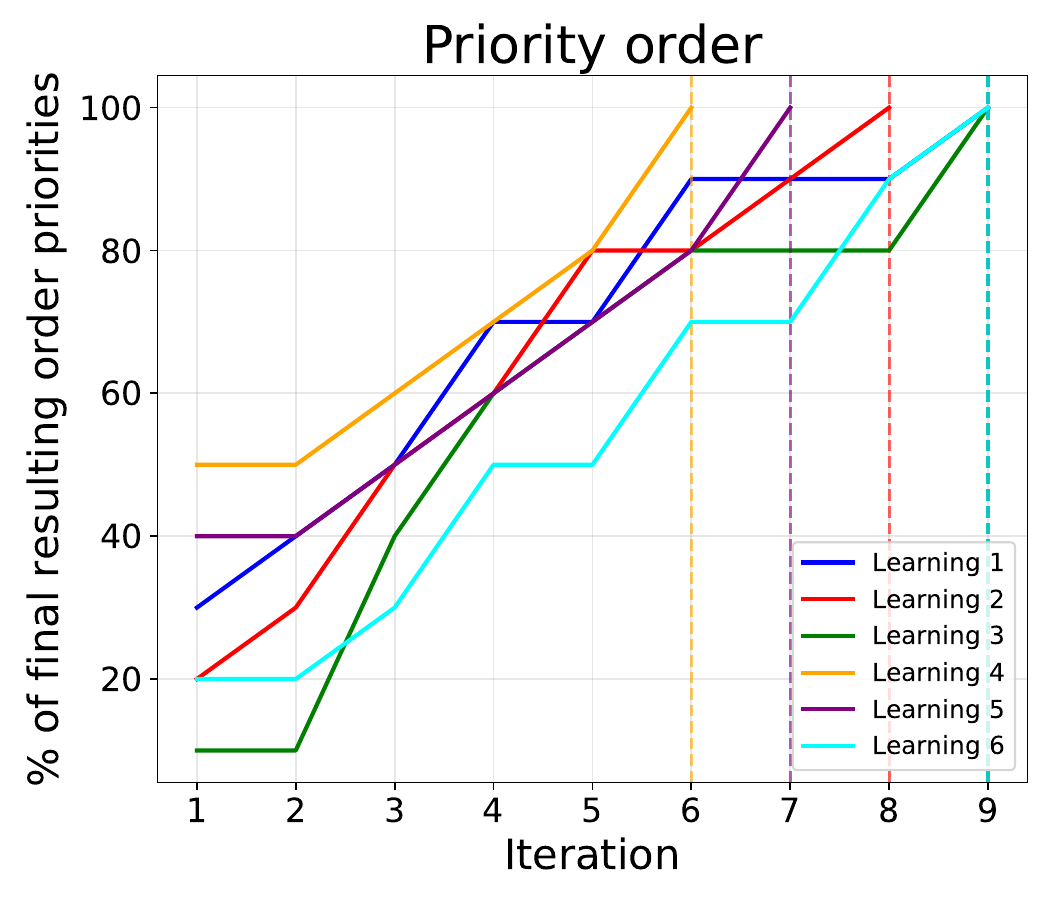}
        \caption{Example with the result of the \% of individuals in a generation that have the same priority order as the final solution, in different learning runs. The base only case is reported on the left, in which the order is [OA, IK]. On the right, the result obtained with the right arm + base, in which the order is [OA, IK, Max. Manip., Max MJL].}
        \label{fig: example 1}
    \end{figure}

During priority learning, task parameters $\theta_i$ are fixed to stable baseline values. Across 30 independent learning trials, the algorithm consistently converged to the same dominant order, with only variations in the last two positions due to different cost functions. Fig.~\ref{fig: example 1} shows representative convergence curves. The plots illustrate how the percentage of individuals that match the final priority order increases over successive generations.
For the mobile base alone, convergence occurred in 3--4 generations on average, due to a limited task set (IK and OA). In contrast, when arm-related tasks were included (up to four tasks), convergence required 7--9 generations, depending on the randomized initial population configuration.
\\Across all runs, the most frequently learned order was:
\begin{itemize}
    \item $1^{st}$ Obstacle Avoidance (OA)
    \item $2^{nd}$ Inverse Kinematics (IK)
    \item $3^{rd}$ Maximization of Manipulability
    \item $4^{th}$ Maximization of distance from Mechanical Joint Limits (M.J.L.)
\end{itemize}
This order reflects the high penalty assigned to collisions and the need to preserve controllability while pursuing the desired pose. While the last two tasks occasionally swap, the first two priorities were invariant across all experiments. This stability indicates that the cost-based evolutionary mechanism reliably uncovers the intuitive and safety-driven task structure normally assigned by human experts. Learning parameters are reported in Tab~\ref{tab:prior learning}. 
\begin{table}[h]
    \centering
    \begin{tabular}{|c|c|}
    \hline
        \textbf{Learning parameters} & \textbf{Value} \\\hline
         number of SoTs in population& 10 \\\hline\hline
         \textbf{Genetic operations probabilities} & \textbf{Value} \\\hline
         probability of doing mutation & 0.5 \\\hline
         probability of doing crossover & 0.5 \\\hline\hline
         \textbf{In-operation probabilities} & \textbf{Value} \\\hline
         probability of changing activation flag& 0.3 \\\hline
         probability of changing priority order & 0.5 \\\hline
         probability of taking priority order and flag as they are & 0.2 \\\hline
    \end{tabular}
    \caption{Learning parameters for Genetic Programming algorithm. Those values are found empirically and show convergence in all the cases tested.}
    \label{tab:prior learning}
\end{table}
\subsubsection{Parameters learning}
\begin{figure}[]
    \centering
    \includegraphics[width=0.49\linewidth]{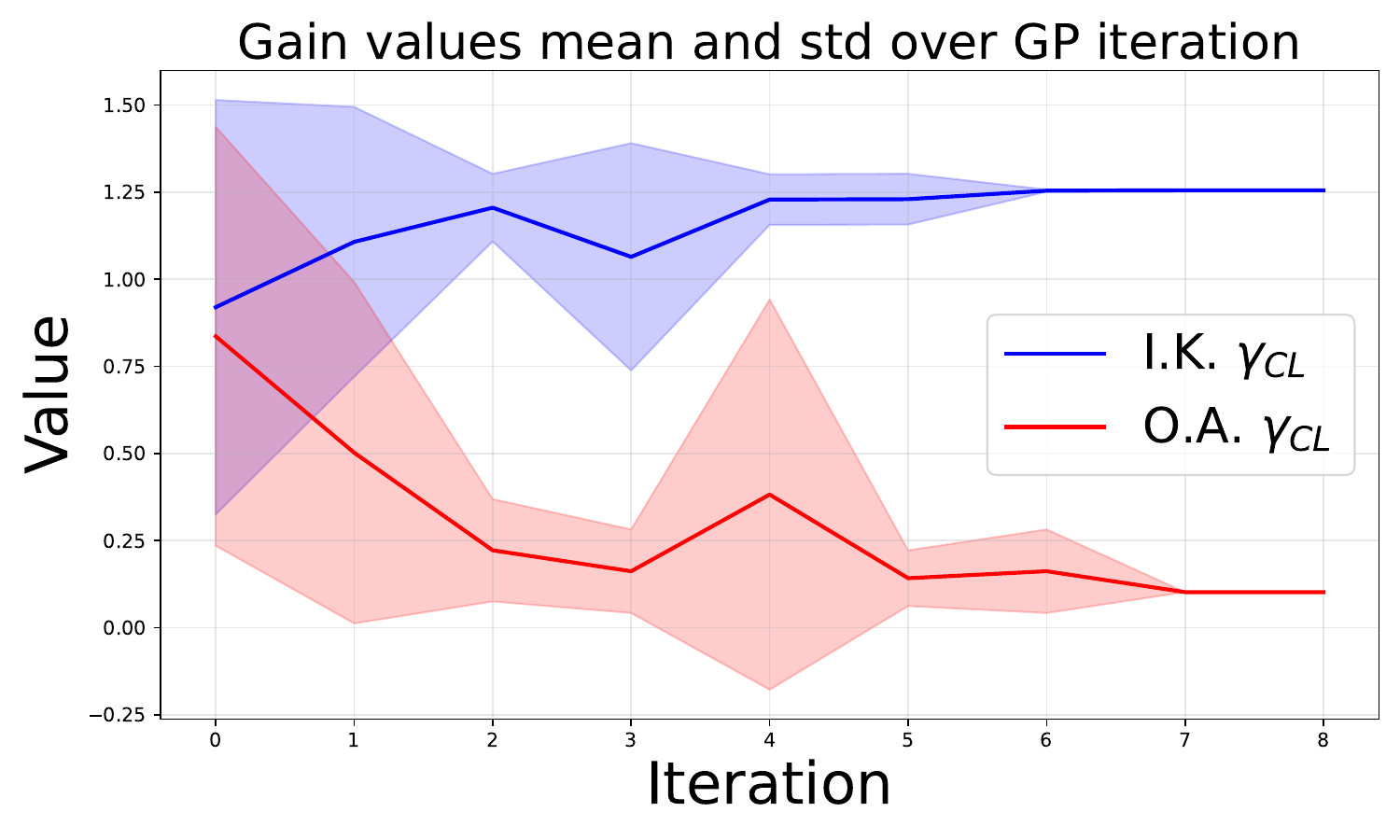}
    \includegraphics[width=0.49\linewidth]{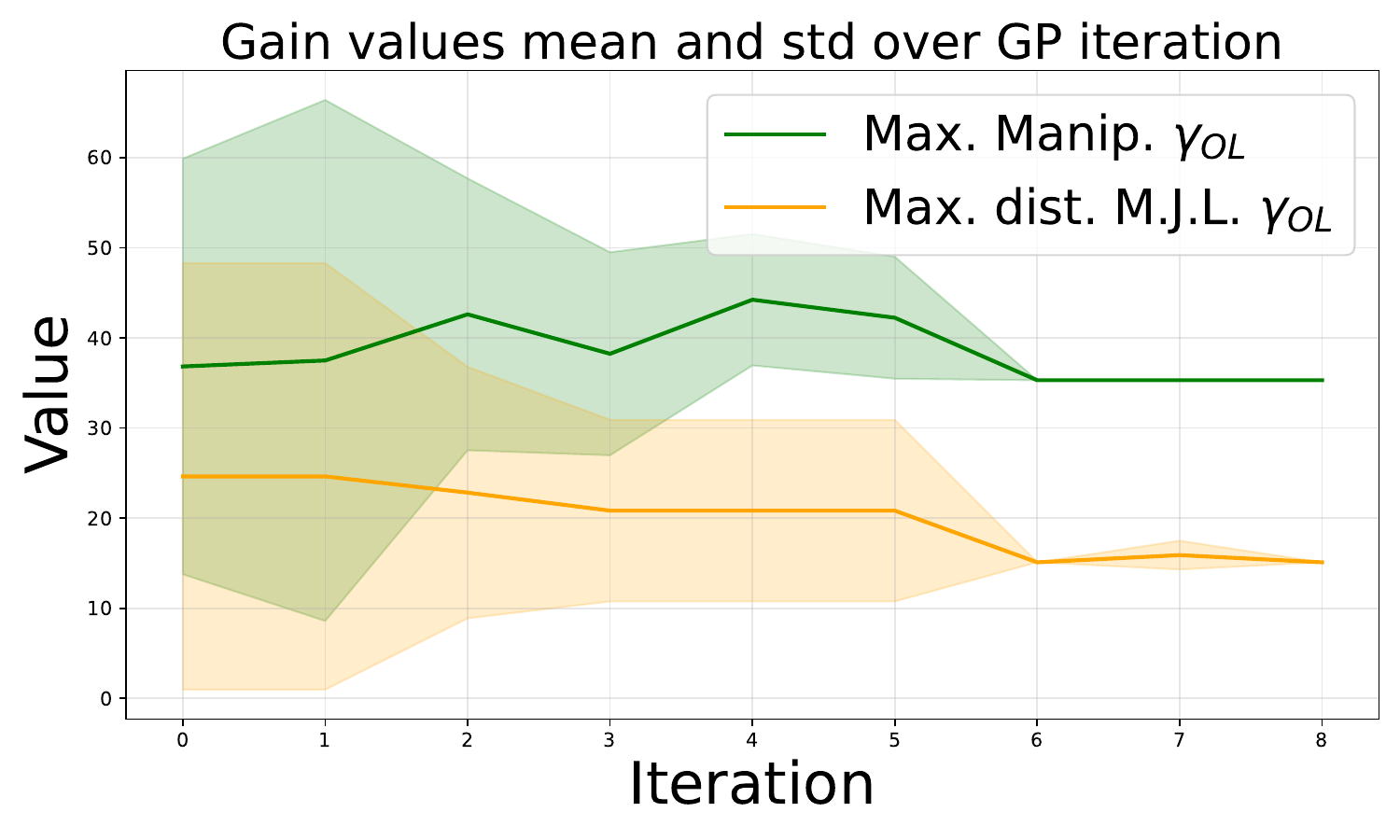}
    \caption{Mean and standard deviation for gains in population after tournament selection for the four different tasks during learning epochs for $C_1$. High standard deviation is due to GP introducing new individuals with different gains, enhancing exploration of the search space.}
    \label{fig:gains bh}
\end{figure}

Once priorities are fixed, parameter learning is performed with all task gains and trajectory times free to vary within predefined safety bounds. The cost function has a greater impact on the optimization of the parameters than on the priority order. Therefore, different cost definitions lead to distinct equilibria. During the parameter optimization run, all SoTs are initialized with the priority order found in the previous step. However, specific cost functions might still swap the task order to refine the structure further.
\\We illustrate this in an example where we evaluate two contrasting cost functions.
\begin{table}[h]
    \centering
    \begin{tabular}{|c|c|}
    \hline
        \textbf{Learning parameter} & \textbf{Value} \\\hline
         number of SoTs in population& 10 \\\hline\hline
         \textbf{Genetic operations probabilities} & \textbf{Value} \\\hline
         probability of doing mutation & 0.5 \\\hline
         probability of doing crossover & 0.5 \\\hline\hline
         \textbf{In-operation probabilities} & \textbf{Value} \\\hline
         probability of changing activation flag& 0.15 \\\hline
         probability of changing a priority& 0.15 \\\hline
         probability of changing a parameter & $\frac{1}{\#_{params.}}-0.3$ \\\hline
    \end{tabular}
    \caption{Learning parameters for Genetic Programming algorithm. Those values are found empirically and showed convergence in all the cases tested.}
    \label{tab:placeholder}
\end{table}
\paragraph{Case 1: Multi-criteria cost (safety + precision + manipulability).}
The first cost
\[
C_1 = 0.4\,c_{\text{prec}} + 0.5\,c_{\text{min.dist}} + 0.1\,c_{\text{max.manip}}.
\]
encourages the robot to maintain safe distances, reach a target pose, and avoid singularities.
\\Over 10 trials, convergence consistently occurred between 6 and 9 generations. The learned SoT (Fig.~\ref{fig:sot_C1}) shows high gains in OA to ensure rapid repulsion from obstacles, intermediate IK gain to guarantee stable but responsive convergence, and a non-zero manipulability gain supporting arm posture optimization.
M.J.L. task is deactivated in 87\% of runs, confirming irrelevance for this cost and resilience of the algorithm to that kind of task, avoiding unnecessary and potentially compromising behaviors. The best SoT found for this cost function is reported in Fig.~\ref{fig:sot_C1} while Fig.~\ref{fig:gains bh} shows the behavior of the tasks' gains during iteration epochs and their convergence toward a unique value.

\begin{figure}[H]
\centering
\begin{tikzpicture}[
    task/.style={
        rectangle, rounded corners,
        black, thick,
        minimum width=6cm,
        inner sep=4pt,
        font=\small,
    },
    active/.style={fill=blue!10},
    inactive/.style={fill=gray!10},
]

\node[task,active,align=center] (t1) {\textbf{1. Obstacle Avoidance } \\ 
{\footnotesize $r_k = 0.102,\;\gamma_{CL}=1.726,\;t_{\text{traj}}=1.385$}};
\node[task,active,align=center, below=2pt of t1] (t2) {\textbf{2. Inverse Kinematics } \\
{\footnotesize $\gamma_{CL}=1.255,\;t_{\text{traj}}=3.609$}};
\node[task,active,align=center, below=2pt of t2] (t3) {\textbf{3. Max Manipulability } \\
{\footnotesize $\gamma_{OL}=35.307$}};
\node[task,inactive,align=center, below=2pt of t3] (t4) {\textbf{4. Max Distance M.J.L. (inactive) } \\
{\footnotesize $\gamma_{OL}=15.001$}};

\end{tikzpicture}
\caption{Best Stack of Task learned for 
$C_1 = 0.4\,c_{\text{prec.}} + 0.5\,c_{\text{min.dist.}} + 0.1\,c_{\text{max.manip.}}$.}
\label{fig:sot_C1}
\end{figure}
\paragraph{Case 2: Speed–precision trade-off.}

The second cost
\[
C_2 = 0.5\,c_{\text{prec}} + 0.5\,c_{\text{time}},
\]
encourages fast convergence without explicit safety measures. Nonetheless, OA remains at the highest priority because collisions introduce a prohibitive penalty.
\\Compared to $C_1$, learned IK gains were, on average, higher, and trajectory times were shorter, illustrating adaptation to a speed-oriented objective. Manipulability and M.J.L. tasks were frequently deactivated (in 70–90\% of trials), reflecting their low contribution to the cost.
The best SoT found for this cost function is reported in Fig.~\ref{fig:sot_C2}.
\begin{figure}[H]
\centering
\begin{tikzpicture}[
    task/.style={
        rectangle, rounded corners,
        black, thick,
        minimum width=6cm,
        inner sep=4pt,
        font=\small,
    },
    active/.style={fill=blue!10},
    inactive/.style={fill=gray!10},
]

\node[task,active,align=center] (t1) {\textbf{1. Obstacle Avoidance } \\ 
{\footnotesize $r_k = 0.477,\;\gamma_{CL}=1.257,\;t_{\text{traj}}=1.030$}};
\node[task,active,align=center, below=2pt of t1] (t2) {\textbf{2. Inverse Kinematics } \\
{\footnotesize $\gamma_{CL}=1.623,\;t_{\text{traj}}=2.641$}};
\node[task,active,align=center, below=2pt of t2] (t3) {\textbf{3. Max Manipulability } \\
{\footnotesize $\gamma_{OL}=8.628$}};
\node[task,inactive,align=center, below=2pt of t3] (t4) {\textbf{4. Max Distance M.J.L. (inactive) } \\
{\footnotesize $\gamma_{OL}=51.074$}};

\end{tikzpicture}
\caption{Best Stack of Task learned for 
$C_2 = 0.5\,c_{\text{prec.}} + 0.5\,c_{\text{time}}$.}
\label{fig:sot_C2}
\end{figure}

\subsubsection{Robustness to non-relevant tasks}

To test robustness, 2–4 distracting tasks were inserted at random positions within the set of available tasks. These tasks did not benefit the mission and occasionally blocked primary tasks if assigned a high priority.
\\Across all trials (30 total), non-relevant tasks were consistently pushed to the last positions or deactivated:
\begin{itemize}
    \item In 100\% of the runs, all distracting tasks were deactivated at the end.
    \item On average, distracting tasks reached the last position within 4–11 generations, depending on their initial location.
\end{itemize}

\begin{figure}[]
        \centering
        \includegraphics[width=0.49\linewidth]{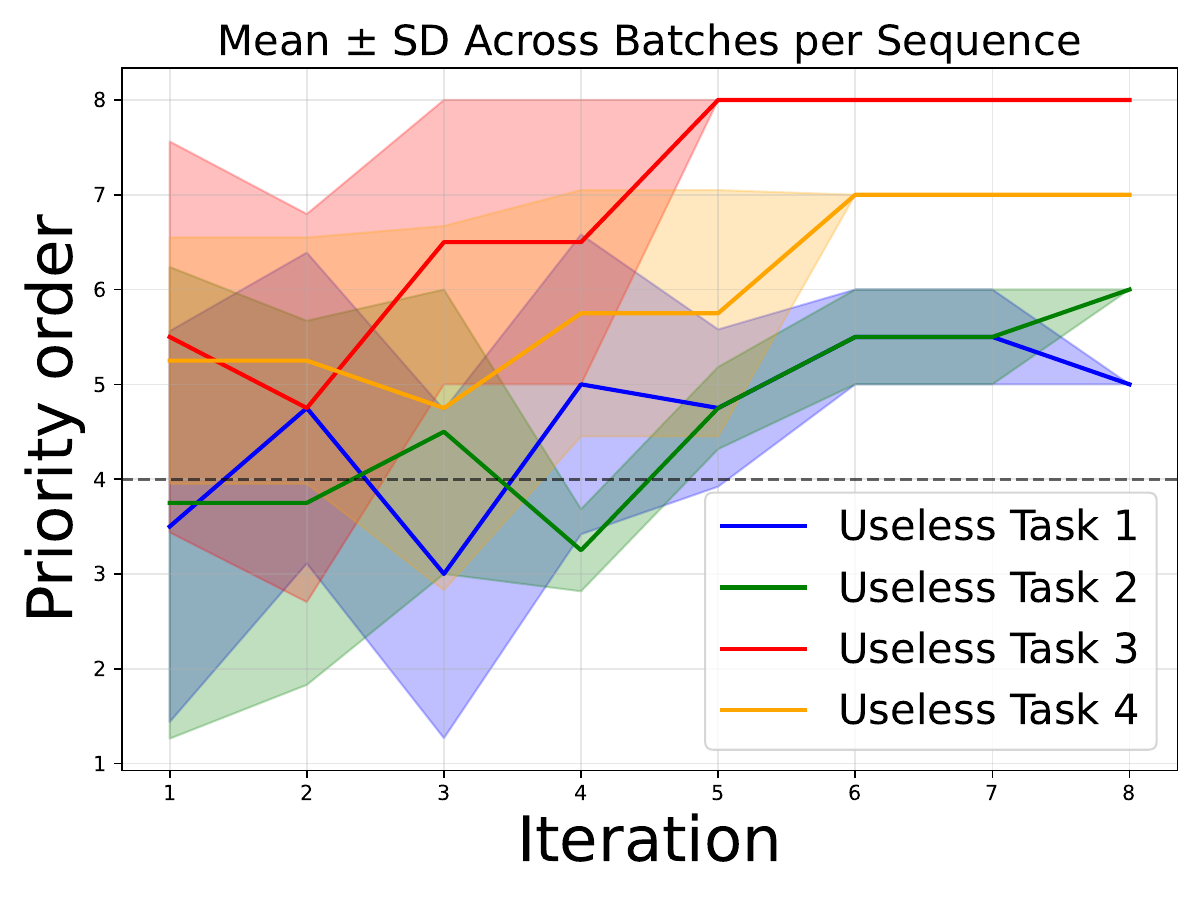}
        \caption{Example of robustness to non-relevant tasks over a learning priority process. After 5 iterations, useless tasks have a higher priority order ($\geq5$) than relevant tasks. After 8 iterations, the order is the same for all SoTs.}
        \label{fig:useless idx}
    \end{figure}

Fig.~\ref{fig:useless idx} illustrates a typical convergence trend for a task initially ranked first.
This shows that the genetic mechanism reliably filters noise and preserves the semantic core of the relevant behaviors.

\begin{figure*}[htbp]
    \centering
    \includegraphics[width=0.24\textwidth]{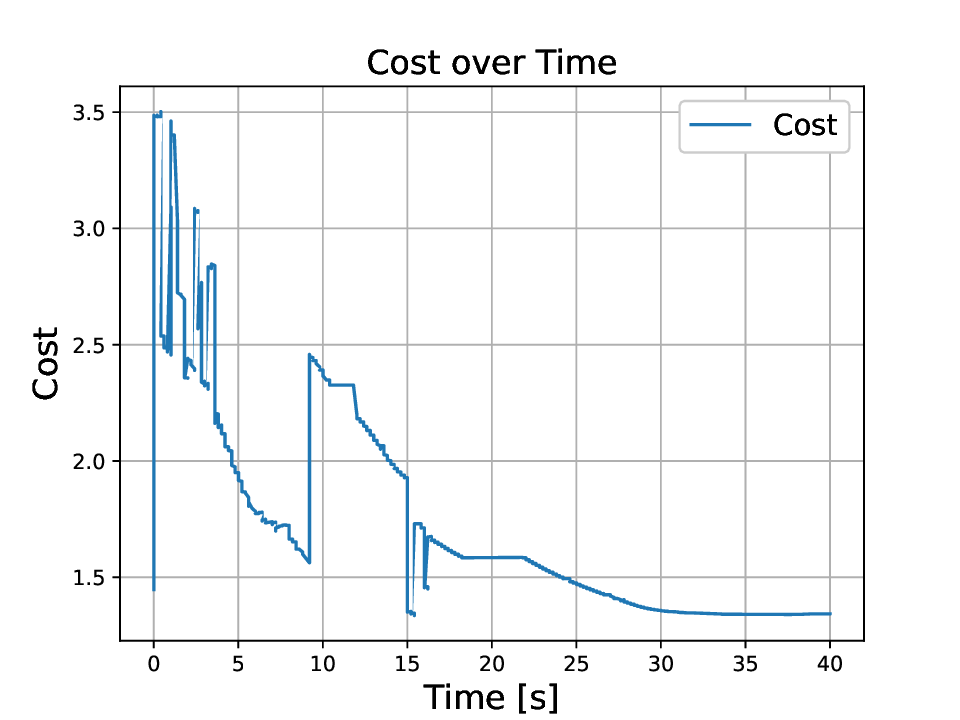}
    \includegraphics[width=0.24\textwidth]{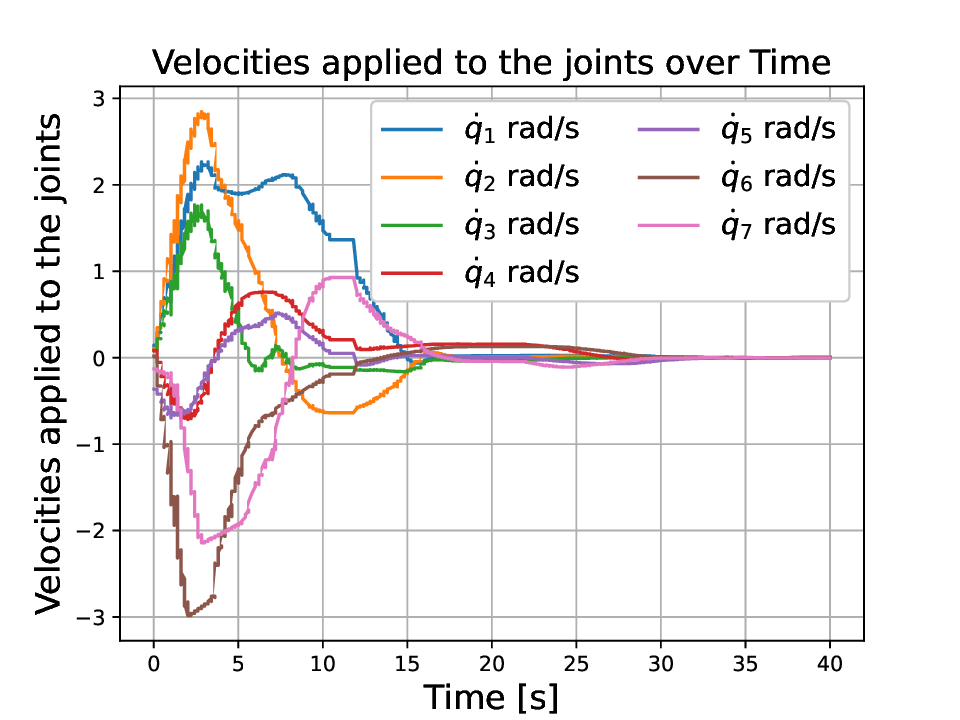}
    \includegraphics[width=0.24\textwidth]{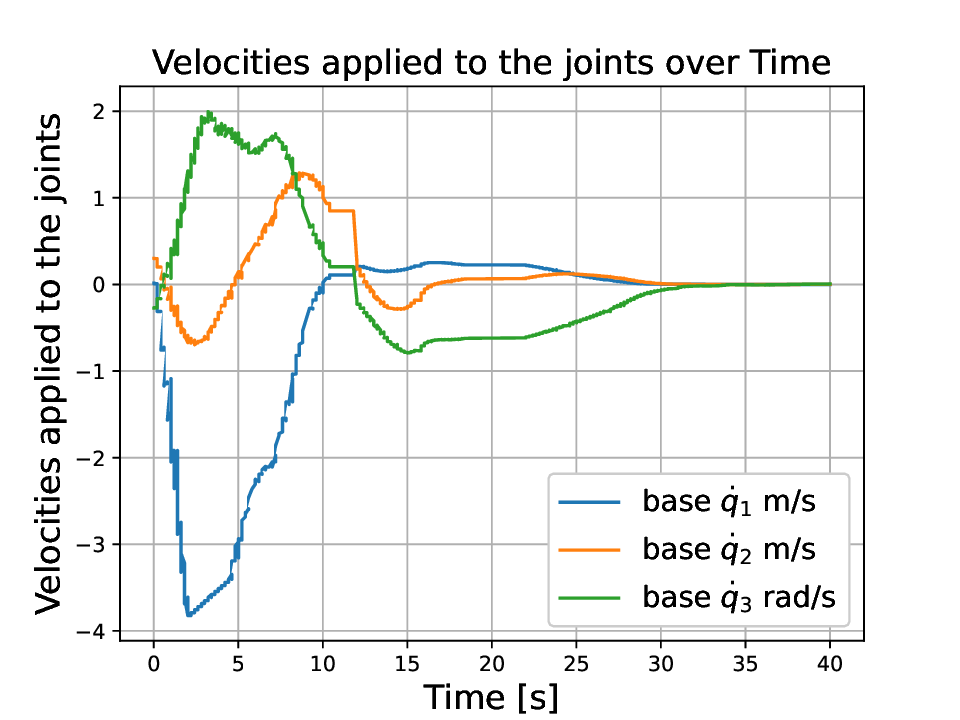}
    \includegraphics[width=0.24\textwidth]{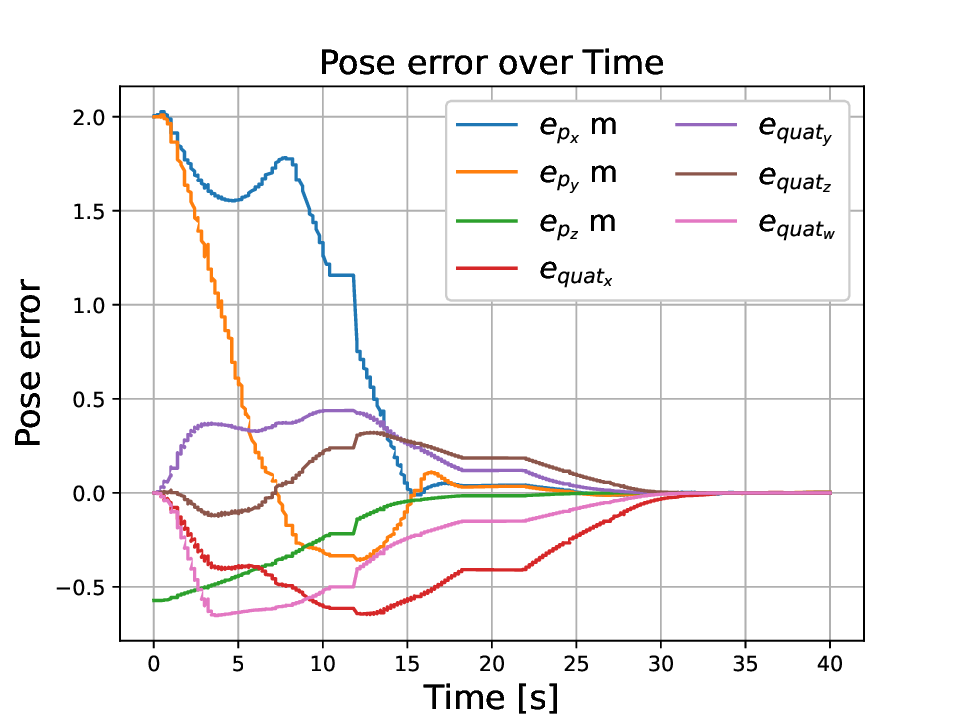}
    \includegraphics[width=0.24\textwidth]{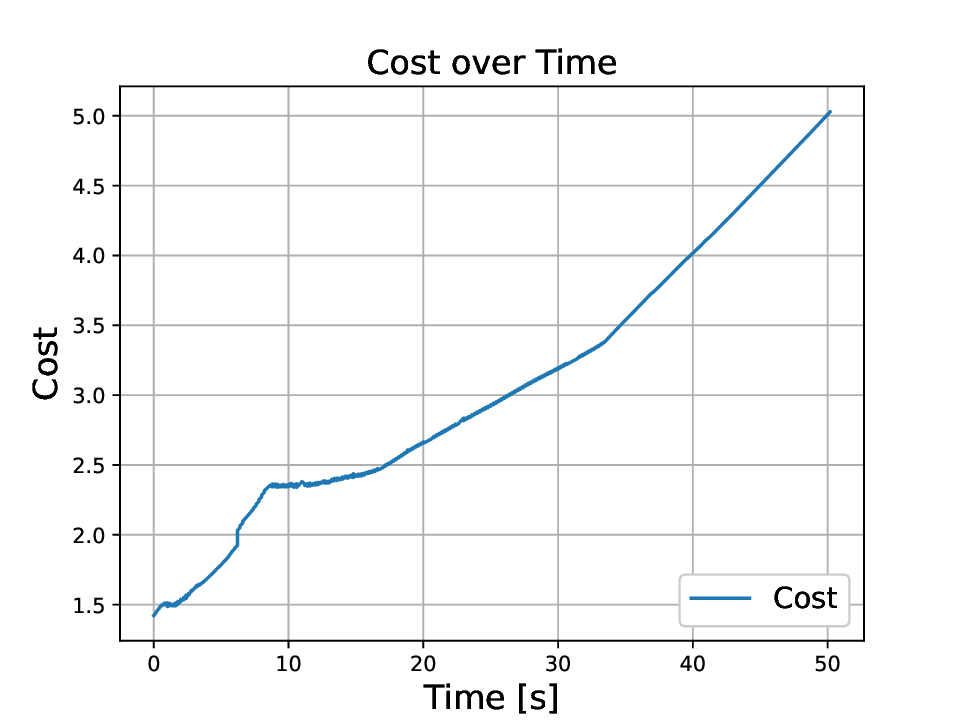}
    \includegraphics[width=0.24\textwidth]{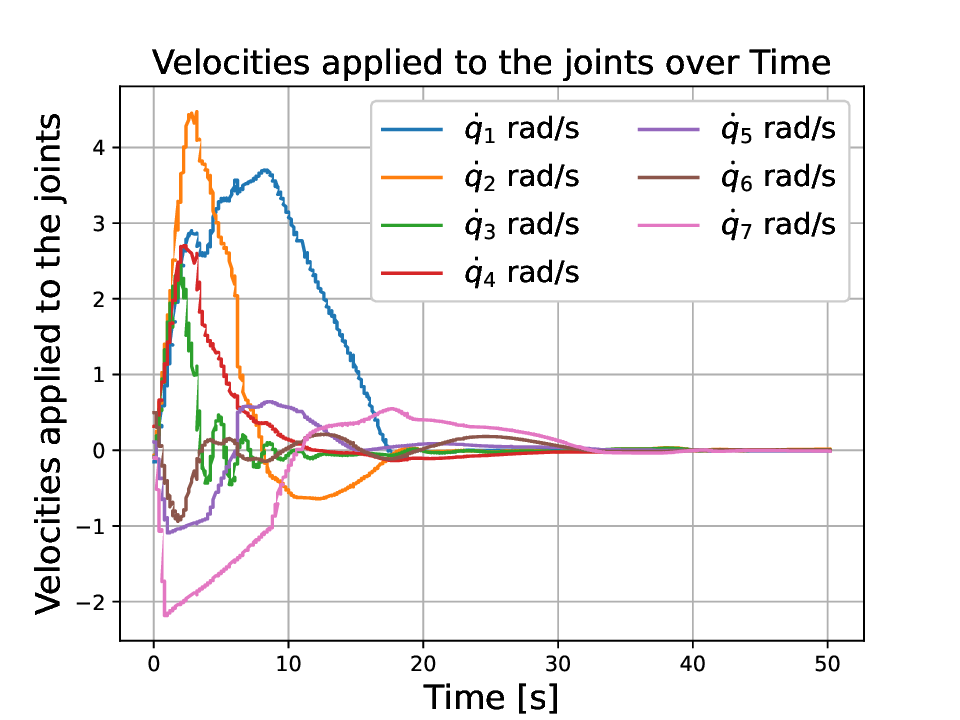}
    \includegraphics[width=0.24\textwidth]{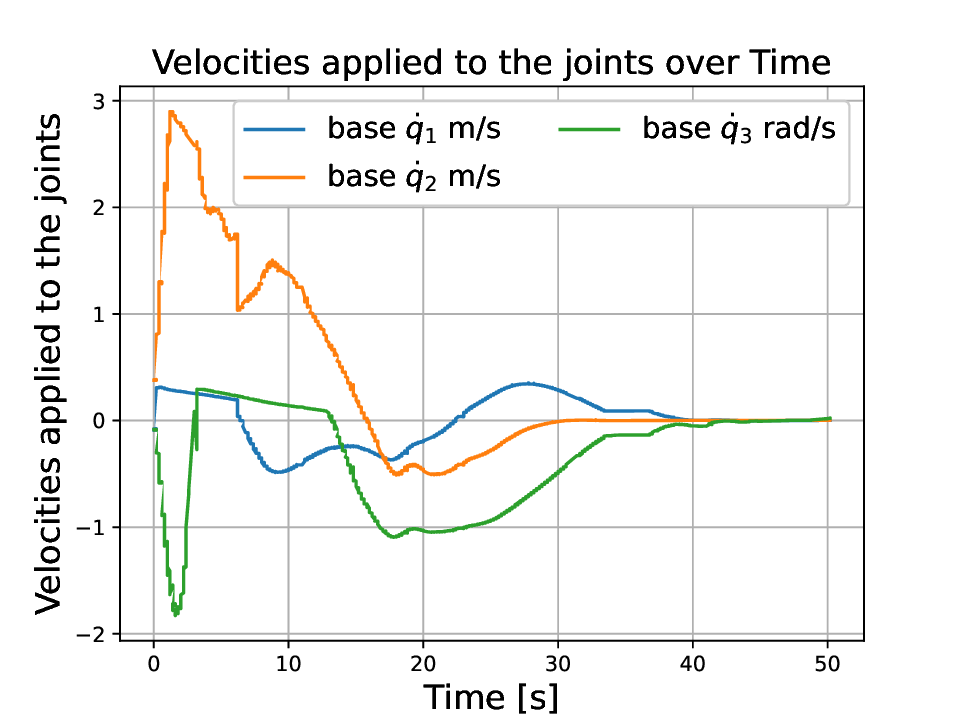}
    \includegraphics[width=0.24\textwidth]{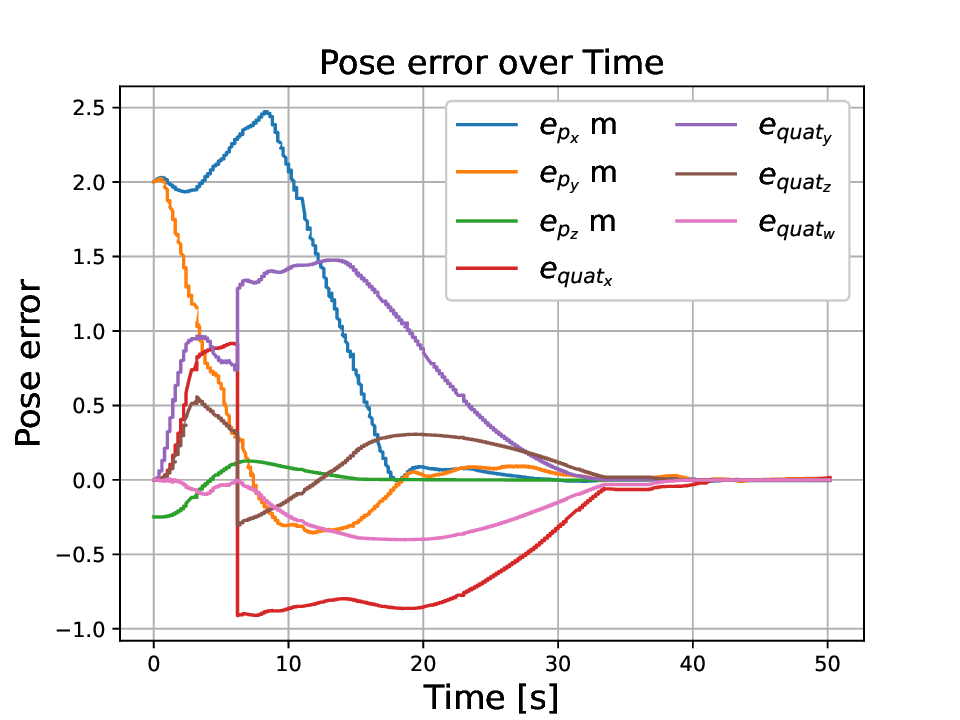}
    \caption{Laboratory results for $C_1$ (top) and $C_2$ (bottom): cost evolution, joint velocities, base velocities, and IK pose error.}
    \label{fig:lab test}
\end{figure*}

\subsection{Laboratory tests}
Learned SoTs were evaluated on the real mobile-YuMi (Fig.~\ref{fig:lab test}) under both static and dynamic conditions.

\paragraph{Transfer to real hardware}
Across all experiments, all learned SoTs were executed without modification.
\\The robot successfully avoided collisions in $100\% $ of static trials and $92\%$ of dynamic trials (in the remaining $8\%$, the robot stopped preemptively due to safety overrides triggered by noisy sensor readings).
\paragraph{Static vs. dynamic environments}
The robot occasionally deviated significantly from the target, prioritizing safety in accordance with learned priorities, behaving safely as expected.
\paragraph{Comparison with simulation}
Minor discrepancies arose due to model inaccuracies and sensor noise: OA activation thresholds were triggered earlier in the real robot, the base motion was smoother in simulation, while the real robot exhibited more conservative adjustments. Despite these differences, the SoTs remained valid and preserved mission success.
\\Finally, while no direct numerical comparison to previous SoT frameworks exists (prior work relies on manually crafted hierarchies), the learned priorities align with those commonly adopted by expert designers (e.g., ~\cite{LowLovelFlexiblePlanning}), supporting the interpretability and practical validity of our automatically learned task structures. In Sec .~\ref{sec: comparison}, comparison with already existing techniques for the design of SoT is addressed.

\subsection{Comparison with Hand-Crafted \& AI-Generated SoT}
\label{sec: comparison}
Traditional Stack-of-Tasks (SoT) controllers rely on expert-defined hierarchies, where priorities and gains are manually selected and tuned for specific robots, environments, as well as for the specific application and mission. Although effective in structured scenarios, these designs exhibit limited ease of use by non-expert users and require substantial retuning when mission objectives or operating conditions change. By contrast, the proposed learning framework automatically derives both task priorities and parameters from simulation, guided solely by a user-defined cost function whose parametrization is facilitated by a GUI.

\paragraph{Hand-crafted SoTs}
Hand-designed SoTs typically assign the highest priority to safety-related tasks, followed by goal-tracking and posture-optimization tasks, consistent with classical designs such as \cite{PickAndPlaceYumi,LowLovelFlexiblePlanning}. In our experiments, the reference hierarchy
\[
[\,\text{OA} \rightarrow \text{IK} \rightarrow \text{Manipulability} \rightarrow \text{M.J.L.}\,]
\]
matched the most frequently learned structure, confirming that the optimizer rediscovered expert-like configurations when the cost emphasized safety and precision. However, manually tuned gains and the hand-crafted hierarchy could not adapt automatically when tasks or objective weights changed, requiring manual fine-tuning.

\paragraph{LLM-generated SoTs}
As a point of comparison, we also evaluate SoTs generated directly by a large language model (LLM), without any subsequent optimization or performance feedback. The LLM is prompted with a natural-language description of the task dictionary, controller parameters, and the high-level cost function to be prioritized, and is asked to propose a complete task hierarchy with associated gains and activation parameters. This setting reflects a common emerging use case in robotics, where LLMs are employed as heuristic planners or design assistants to generate control structures based on semantic knowledge rather than quantitative optimization. ChatGPT-5 was used as the SoTs generator.
\\The learned SoTs reproduced expert-like priorities in all experiments involving safety or precision, while automatically adjusting both priority order and gains to match alternative trade-offs. For example, under the speed-oriented cost~$C_2$, IK gains increased and trajectory durations decreased, whereas under $C_1$, OA gains were elevated and manipulability was promoted to avoid singularities. Unlike LLM-generated SoTs, the learned controllers consistently reflected the cost function. A further advantage is robustness to irrelevant tasks: distracting tasks were reliably deactivated or relegated to low-priority positions, supporting scalability to larger task dictionaries or integration with high-level planners.
\begin{table}[]
    \centering
    \begin{tabular}{|c|c|}
    \hline
        \textbf{Cost function} & \textbf{Generated SoT} \\\hline
        $C_1$ & \makecell{$1^{st}$[OA, True, gain = 2, dist = 0.6 m, t = 4 s] \\ $2^{nd}$[Max. dist. M.J.L., True, gain = 80] \\ $3^{rd}$[IK, True, gain = 1.6, t = 4 s]\\ $4^{th}$[Max. Manip., True, 20]}\\
         \hline
         $C_2$ & \makecell{$1^{st}$[OA, True, gain = 1.0, dist = 0.35 m, t = 4 s] \\ $2^{nd}$[Max. dist. M.J.L., True, gain = 40] \\ $3^{rd}$[IK, True, gain = 2.0, t = 1.5 s]\\ $4^{th}$[Max. Manip., True, 10]}\\
         \hline
    \end{tabular}
    \caption{SoTs generated by LLM (ChatGPT-5) for $C_1$ and $C_2$. In both cases, it is possible to see how the IK task has a lower priority than maximisation of distances from mechanical joint limits. This prevents the robot from reaching the desired pose. In this scenario, expert hand-tuning of the priorities is still required.}
    \label{tab:chat sots}
\end{table}
The LLM output, like the hand-crafted approach, provides only a generated hierarchy, not an optimized one. As shown in Tab.~\ref{fig:lab test}, the LLM-generated priorities were unsuitable for completing the desired objective, preventing the robot from reaching the target pose by subordinating the core IK task. This inability to correctly tune gains and establish a cost-aligned hierarchy quantitatively demonstrates the necessity of the proposed systematic optimization framework over a heuristic generation method, confirming that no existing technique currently solves the full SoT optimization problem.

\subsection{Computational Complexity, Convergence, and Practical Considerations}
The proposed learning framework was executed in \textit{Gazebo} using a full-dynamics model of the mobile-YuMi robot, enabling high-fidelity evaluation of Stack-of-Tasks (SoT) candidates and supporting transfer to hardware. Under this setup, Genetic Programming (GP) operated at near real-time simulation rates, indicating that the approach is feasible for iterative SoT optimization and would incur even lower computational cost when deployed in lighter-weight simulators like MuJoCo~\cite{todorov2012mujoco}, or reduced-order models. The ROS~2-based pipeline further facilitated seamless deployment to the physical robot.
\\GP operates in a mixed discrete--continuous search space, where task ordering and activation flags represent discrete variables, and task gains and trajectory durations define continuous parameters. For a task set of size $K_T$, population size $P$, and $G$ generations, the search domain factorizes as
\begin{equation}
\mathcal{S} = \mathcal{P}(K_T) \times \Theta(K_T),
\end{equation}
with up to $K_T!$ priority permutations and $2^{K_T}$ activation patterns. Exhaustive exploration of such a space is intractable; however, GP has demonstrated strong empirical performance in similarly structured hybrid domains \cite{poli2008field, vladislavleva2009order, langdon2013foundations}. The overall computational cost is dominated by simulation-based evaluation,
\begin{equation}
\mathcal{O}(P \cdot G \cdot T_{\text{sim}}),
\end{equation}
consistent with observations in simulation-driven evolutionary robotics \cite{bongard2013evolutionary, mouret2013illumination}. In our experiments, convergence was achieved within 5--12 generations, corresponding to total runtimes of 1.5--3\,h for populations of 10 individuals with episode durations of approximately 40\,s plus the time needed to re-initialize the simulation.
\\Convergence behavior aligns with established results in GP: rapid progression is observed when evaluation noise is low, genetic diversity is preserved, and intron-based inactive tasks mitigate destructive crossover \cite{nordin1995explicit, stanley2002evolving}. Priority order converged within a few generations for base-only tasks and more for full robot configurations as the number is higher, while parameter tuning required an additional number of generations. This matches the convergence characteristics reported in \cite{poli2008field, vladislavleva2009order} for mixed discrete--continuous domains.
\\Because SoT evaluation in simulation is deterministic and episodic, the resulting cost landscape is sufficiently smooth for GP to consistently rediscover expert-like task hierarchies. Activation flags also act as effective bloat regulators, reducing the functional search dimension and improving robustness. Although GP provides no formal guarantee of global optimality, the observed efficiency and reliability across all experiments support its suitability for learning SoT structures in high-dimensional redundant robotic systems.
\section{Conclusion and Future Work}
\label{Conclusion and future works}

This work presented an automatic framework for learning complete Stack-of-Tasks (SoT) controllers, including task priorities, activation logic, and control parameters, directly from a user-defined cost function. By using Genetic Programming within a simulation-based episodic evaluation, the method discovers interpretable and effective task hierarchies for redundant robotic systems without manual tuning.
\\The learned SoTs consistently reflected safety and performance objectives and demonstrated strong generalization across operating conditions. Extensive experiments in Gazebo and on a dual-arm mobile robot (mobile-YuMi) confirmed reliable sim-to-real transfer, safe task execution, and robustness to irrelevant tasks. These results highlight the potential of evolutionary structure learning as a scalable alternative to manually engineered SoT pipelines.
\\Despite these advantages, several limitations remain. Training relies on simulation fidelity and incurs non-negligible computational time; discrepancies due to sensing noise or dynamic uncertainty can affect real-world behavior. Additionally, richer task interactions and more expressive objective functions may require enhanced optimization strategies.
\\Future work will investigate: richer and automatically generated cost structures, including language- or demonstration-based specification; integration with global navigation and planning layers~\cite{Nav2Tool}; extensions to multi-robot and heterogeneous systems; and online or adaptive refinement of gains and priorities, potentially combining GP with gradient-based methods. Advances along these directions will further improve the robustness and autonomy of SoT-based control architectures.
\\Overall, the proposed framework provides a generalizable and user-oriented approach to redundancy resolution, combining the interpretability of classical SoTs with the flexibility of data-driven optimization. A further natural direction for future work is the inclusion of systematic user studies, once the framework is more mature, to quantitatively assess usability, transparency, and effectiveness for non-expert users in realistic applications.

\renewcommand*{\bibfont}{\footnotesize}
\printbibliography
\end{document}